%% file: ShapeFromX.tex
\documentclass[journal]{IEEEtran}
\usepackage{booktabs} 

\usepackage{amsmath}
\usepackage{graphicx}
\usepackage{times}
\usepackage{caption2}

\usepackage{amsmath}

\usepackage[bookmarks,backref=true,linkcolor=black]{hyperref}
\hypersetup{
  pdfauthor = {},
  pdftitle = {},
  pdfsubject = {},
  pdfkeywords = {},
  colorlinks=true,
  linkcolor= black,
  citecolor= black,
  pageanchor=true,
  urlcolor = black,
  plainpages = false,
  linktocpage
}
\usepackage{color}

\usepackage{bm}
\usepackage{amsfonts}
\usepackage{amssymb}
\usepackage[mathscr]{euscript}
\usepackage{comment}
\usepackage{bm}
\usepackage{subfigure}
\usepackage{algorithm}
\usepackage{algorithmic}
\usepackage{multirow}
\usepackage{wrapfig}
\usepackage{booktabs}
\usepackage{threeparttable}
\usepackage[english]{babel}
\usepackage{microtype}
\usepackage{url}
\usepackage{xcolor}

\newcommand{\bfA}{\mathbf{A}}
\newcommand{\bfb}{\mathbf{b}}

\newcommand{\bfI}{\mathbf{I}}

\newcommand{\bfn}{\mathbf{n}}
\newcommand{\bfp}{\mathbf{p}}

\newcommand{\bfq}{\mathbf{q}}
\newcommand{\bfpi}{\mathbf{\Pi}}
\newcommand{\bfr}{\mathbf{r}}

\newcommand{\bft}{\mathbf{t}}

\newcommand{\bfR}{\mathbf{R}}
\newcommand{\bfalpha}{\bm{\alpha}}

\newcommand{\txalb}{\textrm{alb}}

\newcommand{\pixel}{m}
\newcommand{\pixels}{\mathcal{F}}

\newcommand{\edit}{\textcolor{black}}
\newcommand{\edittip}{\textcolor{black}}

\begin{document}
\title{3D Face From X: Learning Face Shape from Diverse Sources}\author{Yudong~Guo, \,\,\,\,Lin~Cai, \,\,\,\,Juyong~Zhang$^\dagger$
\thanks{Yudong~Guo and Juyong~Zhang are with School of Mathematical Sciences, University of Science and Technology of China.}
\thanks{Lin Cai is with Beijing Dilusense Technology Corporation.}
\thanks{$^\dagger$Corresponding author. Email: {\texttt{juyong@ustc.edu.cn}}.}
}

\maketitle

\IEEEtitleabstractindextext{%
\begin{abstract}
We present a novel method to jointly learn a 3D face parametric model and 3D face reconstruction from diverse sources. Previous methods usually learn 3D face modeling from one kind of source, such as scanned data or in-the-wild images. Although 3D scanned data contain accurate geometric information of face shapes, the capture system is expensive and such datasets usually contain a small number of subjects. On the other hand, in-the-wild face images are easily obtained and there are a large number of facial images. However, facial images do not contain explicit geometric information. In this paper, we propose a method to learn a unified face model from diverse sources. Besides scanned face data and face images, we also utilize a large number of RGB-D images captured with an iPhone X to bridge the gap between the two sources. Experimental results demonstrate that with training data from more sources, we can learn a more powerful face model.
\end{abstract}

}

\IEEEpeerreviewmaketitle

\IEEEdisplaynontitleabstractindextext

%
\IEEEpeerreviewmaketitle

\input{Introduction}

\input{related_work}
\input{method}
\input{results}
\input{conclusion}

{\small
\bibliographystyle{ieee}
\bibliography{ShapeFromX}
}

\end{document}

%% file: Introduction.tex
 \section{Introduction}
 \begin{figure*}
 	\centering
 	\includegraphics[width=1\textwidth]{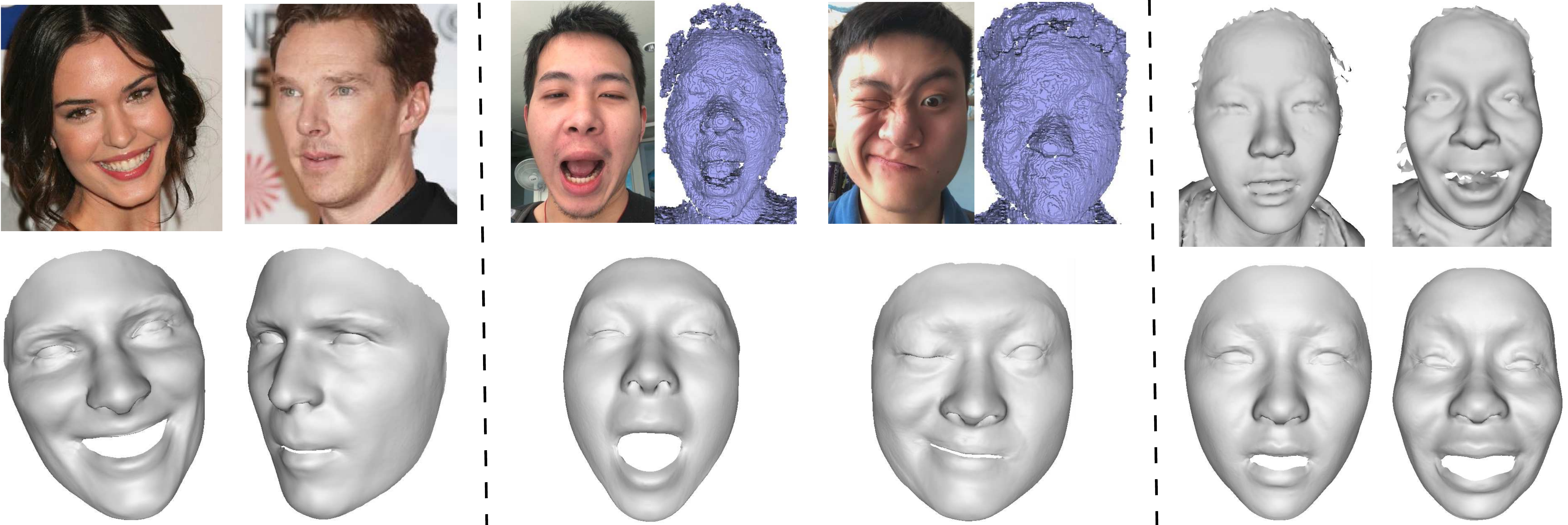}
 	\edit{\caption{Several 3D face reconstruction results from diverse sources. On the top are different inputs from RGB, RGB-D and raw scan data respectively, and on the bottom are corresponding reconstruction results. Our method is the first deep learning based 3D facial performance capture work with inputs from diverse sources.}}
 	\label{fig:system}
 \end{figure*}
 This paper considers the problem of jointly learning a nonlinear 3D Face Morphable Model (3DMM) and 3D face reconstruction with training data from three different sources: scanned 3D face, RGB-D images, and RGB images. 3D face modeling from scanned 3D face dataset or face modeling and capture from RGB images have been well studied in recent years. Since Blanz et al.~\cite{blanz1999face} build a PCA model with 200 scanned 3D faces, there have been several works that build a parametric model from scanned 3D faces~\cite{cao2014facewarehouse,li2017learning}. However, 3D scanned faces are usually expensive to obtain, and the training data usually contain a small number of subjects. On the other hand, in-the-wild RGB images are very easy to obtain and there are many works to learn face modeling and face reconstruction on such dataset~\cite{richardson2016learning,tewari17MoFA,tewari18FaceModel,Guo20193DFace}. However, compared to scanned data, RGB images do not contain geometric information, and it is an ill-posed problem to infer geometry from RGB images. Thus these works have to use facial priors for regularization.
 
 Although 3D face modeling from RGB input have been well studied in recent years, real-time RGB based methods still have difficulty to achieve high-quality and robust facial performance capture due to limited information from the input RGB images. For example, RGB-based methods only have appearance information, and they cannot perform well or even fail in a dark environment. RGB-D sensors, on the other hand, provide not only appearance information but also depth information that is robust to illumination changes and occlusions. With the help of additional depth information, RGB-D based methods could achieve better and more robust performance compared to RGB-based methods. Hence RGB-D based capturing has recently received increasing attention, especially with consumer-grade depth cameras becoming popular. For example, some newly released smartphones like Apple's iPhone X are equipped with an RGB-D camera, which stimulates the call for more efficient and pragmatic RGB-D face capturing solutions on mobile devices.

 To bridge the gap between the reconstruction results of different sources and take advantage of all types of data, we propose to use datasets from all kinds of sources to train a unified face model. Our work is inspired by the great success of deep learning-based technology for face modeling and 3D face reconstruction from one kind of source. For example, many deep learning-based methods have been proposed for 3D face reconstruction based on RGB inputs. \cite{richardson2016learning} and \cite{Guo20193DFace} propose a coarse-to-fine network architecture where a coarse-scale CNN is trained to regress the parameters of a parametric face model and a fine-scale CNN is trained to recover the face detail. \cite{tewari17MoFA} proposes an unsupervised CNN structure including a CNN encoder and a fixed rendering module as the decoder and trains the structure end-to-end for 3D face reconstruction from RGB input. \cite{tewari18FaceModel} further improves the reconstruction accuracy by learning the corrective basis from sparsely labeled in-the-wild images. Very recently, \cite{liu20193d} presents a deep learning-based method to learn face modeling from diverse raw scan data, in which a PointNet~\cite{qi2017pointnet} is used as feature encoder. \edittip{Similar to 3DMM [6], nonlinear 3DMM [65] and FLAME [37], we learn a low-dimensional representation for face modeling. Low-dimensional models can reconstruct global facial structure while can not recover fine-scale details due to the limited representation ability of the low-dimensional parametric model. The advantage of our method over existing low-dimensional models is to jointly learn a powerful face model and 3D face reconstruction from diverse sources.}

 It is however non-trivial to directly adopt existing methods to learn 3D face from diverse sources. There are two challenges that make it difficult to learn a unified face model from diverse sources. The first challenge is the gap between different sources: the scanned data live in 3D space while the facial images live in 2D space, which makes it difficult to handle different sources with the same network architecture. The second challenge is the lack of dense correspondence. As pointed out in~\cite{liu20193d}, dense correspondence is crucial and challenging for face modeling from different databases. In our case, the problem is more challenging since our data are from different domains.
 
 We adopt several strategies to address these challenges. To bridge the gap between the two sources, we utilize a large number of RGB-D data, which contain information from both types while live in the same domain. For domain consistency, we render scanned 3D face to depth maps and use a three-branches CNN to encode data from different sources to geometric parameters. To make different sources benefit each other, we use cross-domain supervision losses to make the dataset with weak signals benefit from data with stronger supervision signals. Specifically, to make the RGB branch benefit from additional supervision signal from the RGB-D dataset, we feed the RGB images from RGB-D data to the RGB branch and use the depth information as supervision signal. To make the RGB-D branch benefit from more accurate and complete supervision signal from the scan dataset, we render the scan data to RGB-D images with artificial speckle noise by simulating the capturing process of depth cameras. And then feed the synthesized RGB-D images to the RGB-D branch while using original point clouds from scan data as supervision signals.
 
 To tackle the second challenge of building dense correspondence, we first use a fixed face model by fixing the geometry decoder and only updating the encoder network to fit different sources. For scan and RGB-D data, we build dense correspondence between face model and input data by finding the closest point in the input point cloud. For RGB data, we build correspondence by projection and rasterization.  We then update the face model with fixed correspondence obtained in the last step. The two steps are taken iteratively. To sum up, the paper has the following contributions:
 
 \begin{itemize}
 	\item We propose a three-branches CNN learning framework to jointly learn face modeling and 3D face reconstruction with training data from three different sources. \edittip{To the best of our knowledge, this is the first method to learn a unified 3D face model from diverse sources.} Experiments demonstrate that with training data from more sources, we can learn a more powerful face model.
 	
 	\item We bridge the gap of supervision signals from different sources by using novel cross-domain supervision losses to make dataset with weak signals benefit from data with stronger supervision signals. \edittip{Experimental results demonstrate that with additional supervision from RGB-D data during training, our model can infer more accurate geometry for RGB inputs, and that better supervision signals from scan data lead to more accurate reconstruction for RGB-D inputs.}
 	
 	\item To bridge the gap between the scanned face dataset and RGB images dataset, we collect a large scale RGB-D dataset from a variety of identities with different facial expressions, poses, and lighting conditions.~\footnote{Our constructed RGB-D facial dataset and the reconstructed mesh of each frame will be publicly available.} 
 	
 \end{itemize}
 
 We show several reconstruction results of our method on different types of input data in Fig.~\ref{fig:system}. It shows that our method can handle diverse sources well within a unified face model. The qualitative and quantitative comparison with the state-of-the-art RGB and RGB-D based methods in Section~\ref{sec:result} demonstrate that our method performs well on both types of data. We show more results and the learned face model in the supplementary video.

%% file: related_work.tex
\section{Related Work}

\noindent\textbf{3D Face Shape Models.} \edittip{Human faces have similar global characteristics such as the location of eyes, nose and mouth. Based on this prior, some researchers propose to build low-dimensional face models to reduce the reconstruction problem into searching within the parameter space.} Since Blanz and Vetter~\cite{blanz1999face} propose to build a 3D Morphable Model (3DMM) to represent a
textured 3D face with principal components analysis (PCA), low-dimensional face models have been widely used for 3D face representation. Among them, Paysan et al.~\cite{paysan20093d} build the publicly available Basel Face Model (BFM2009) with 200 face scans in a neutral expression. Booth et al.~\cite{booth2018large} utilize neutral scans of $9663$ subjects to improve the representation ability of 3DMM. To also cover facial expressions, Vlasic et al.~\cite{tog-VlasicBPP05} adopt a multilinear face model representation that jointly parameterizes the identity and expression variations. Cao et al.~\cite{cao2014facewarehouse} use a similar representation to build a multilinear 3D face model from $3000$ scans of 150 subjects each with 20 different expressions. Li et al.~\cite{li2017learning} build the FLAME model that additionally represents head rotation and yaw motion with linear blend skinning. 

\edittip{Besides above traditional statistical method for 3D face modeling from scan data, recently there are also some works that apply CNN to learn powerful face shape models from RGB images.} Tran et al.~\cite{tran2018nonlinear} propose to use a multilayer perceptron (MLP) to build a nonlinear form of 3DMM and learn the model with in-the-wild images. \cite{tran2019towards} improves the work by extending to consider facial geometric details. Sengupta et al.~\cite{sfsnetSengupta18} propose to jointly learn the decomposition of shape, albedo and lighting conditions of a face from a collection of RGB images.

\noindent\textbf{\edittip{Optimization-based Facial Performance Capture from RGB Inputs.}} Monocular RGB images or videos can be easily obtained and many methods have been proposed to reconstruct faces from monocular RGB video. \edittip{A popular way is to fit the input images using a low-dimensional parametric face model such as 3DMM with analysis-by-synthesis optimization strategy}. The reconstructed model can be further refined by recovering 3D shape from shading variation~\cite{Kemelmacher-ShlizermanS11,Kemelmacher-ShlizermanB11}. For instance, in \cite{GVWT13},  a template is deformed to a 3D model created by a binocular stereo approach. In \cite{ValgaertsWBST12}, a blendshape model is built and fitted to a monocular video off-line, and then the surface detail is added by shading-based shape refinement under general lighting conditions. \cite{shi2014automatic} uses a similar approach and refines the results by iteratively optimizing the large-scale facial geometry and the fine-scale facial detail. \cite{ichim2015dynamic} proposes to create fully rigged, personalized 3D facial avatars from a video captured from a hand-held camera. The method performs the tracking and reconstruction computation on PC though the input images are captured by mobile devices.
\cite{garrido2016reconstruction} fits a 3D face using a multi-layer approach and extracts a high-fidelity parameterized 3D face rig that contains a generative wrinkle formation model capturing the person-specific idiosyncrasies. As the low-rank parametric face model limits its expressiveness and ability to capture fine details, \cite{SuwajanakornKS14} proposes to derive a person-specific face model from all available images, and each frame of the video is reconstructed by a novel 3D optical flow approach coupled with shading cues. In general, these methods work off-line and are not suitable for real-time 3D facial performance capture. \cite{thies2016face2face} presents a method to jointly fit a parametric model for identity, expression and skin reflectance to the input color, which achieves real-time 3D face tracking and facial reenactment with GPU acceleration.

\noindent\textbf{\edittip{Learning-based 3D Face Reconstruction from RGB Inputs.}}
\edittip{To achieve real-time capturing of 3D faces, many methods utilize machine learning to shift complex optimization to the off-line training stage.} \cite{cao2014displaced} presents a learning-based regression approach to fit a generic identity and expression model to an RGB face video on the fly. \cite{WengCHZ14} extends it to run on mobile devices at real-time frame rates by directly regressing the head poses and expression coefficients in one-step. The approach is also extended to include the learning of fine-scale facial wrinkles~\cite{cao2015real}.

With the powerfulness of convolution neural networks, various deep learning-based methods have been developed for 3D face reconstruction, face alignment, face recognition and dense facial correspondences from RGB images. Examples are \cite{JourablooL17,BhagavatulaZLS17,zhu2016face,KimZTTRT17,corr-LiZZ16c,tran2016regressing,Jin2017FaceAI,GulerTASZK17,YuSLCL17,feng2018joint}. In these methods, 3D Morphable Model (3DMM)~\cite{blanz1999face,RomdhaniV05,booth2018large,luthi2017gaussian} is used to represent 3D faces and CNN is applied to learn the 3DMM and pose parameters. \edittip{Most of these methods train the CNN in a fully supervised manner. \cite{richardson20163d} presents a method for 3D face reconstruction from an RGB image by training a CNN with synthetic data to learn the 3DMM parameters and \cite{richardson2016learning} further trains a fine-scale CNN to regress per-pixel depth displacement for detailed geometry.} \cite{laine2017production} proposes to directly learn the vertex positions from RGB images by building an initial PCA basis. \cite{Guo20193DFace} proposes a fully supervised deep learning framework for reconstructing a detailed 3D face from monocular RGB  video in real time. \cite{JacksonBAT17} proposes to regress volumes with CNN for a single face image directly. \edittip{One problem of these supervised methods is the gap between the synthesized images and real-captured images. To tackle this problem, unsupervised~\cite{tewari17MoFA} and weakly-supervised~\cite{tewari18FaceModel,deng2019accurate} methods have been proposed to train CNNs with real-captured images by using an analysis-by-synthesis energy function.} While these CNN-based methods for 3D face reconstruction from RGB images can achieve impressive real-time performance, they have inherent limitations due to the RGB inputs: easily affected by illumination, unable to handle images captured in a dark environment.

\noindent\textbf{\edittip{Optimization-based Facial Performance Capture from RGB-D Inputs.}}
Compared to RGB inputs, RGB-D inputs provide additional depth information that benefits more robust facial performance capture with better quality. In fact, Weise et al.~\cite{weise2011realtime} developed a facial performance capture system that combines 3D and 2D non-rigid registration in optimization and achieves real-time robust 3D face tracking. However, this system needs to build an accurate 3D expression blendshape by scanning and processing a predefined set of facial expressions in advance. \cite{li2013realtime} describes a system that only requires pre-building a neutral face model, generates initial blendshapes using deformation transfer~\cite{sumner2004deformation}, and then trains PCA-based correctives for the blendshapes during tracking with examples obtained from per-vertex Laplacian deformations. \cite{bouaziz2013online} introduces a calibration-free system that requires no user-specific preprocessing by jointly solving for a detailed 3D expression model of the user and the corresponding dynamic tracking parameters via an optimization procedure.  \cite{chen2013accurate} proposes an approach that incrementally deforms a 3D template mesh model to best match the input RGB-D data and facial landmarks detected from the single RGB-D image. \cite{cvpr-HsiehMYL15} presents a real-time facial tracking system in unconstrained settings using a consumer-level RGB-D sensor. The system personalizes the tracking model on-the-fly by progressively refining the user's identity, expressions, and texture with reliable samples and temporal filtering.

To generate high-quality 3D face models from RGB-D data, \cite{LiXC0K15} proposes an approach consisting of two stages: offline construction of personalized wrinkled blendshape and online 3D facial performance capturing. \cite{thies2015realtime} presents a method that reconstructs high-quality facial performance of each actor by jointly optimizing the unknown head pose, identity parameters, facial expression parameters, face albedo values, and the illumination. The method focuses on photo-realistic capture and re-rendering of facial templates to achieve expression transfer between real actors. Different from fitting the RGB-D data to the parametric model, \cite{LiangKS14} proposes to divide the input depth frame into several semantic regions and search for the best matching shape for each region in the database. The reconstructed 3D mesh is constructed by combining the input depth frame with the matched shapes in the database. \cite{NewcombeFS15} proposes a direct approach that reconstructs and tracks the 3D face without templates or prior models.

%% file: method.tex
\section{Proposed Method}
\label{sec:render}
This section first introduces face representation and rendering procedure used in our method, and then presents the architecture of our three-branches CNN.

\subsection{Face Representation}
The face is basically represented as a vector $\bfp = [\bfp^T_1, \bfp^T_2, \cdots, \bfp^T_{n}]^T\in \mathcal{R}^{3n}$ via a mesh of $n$ vertices $v_i$ ($i = 1, 2, \ldots, n$) with fixed connectivity, where $\bfp_i$ denotes the position of vertex $v_i$. We use a nonlinear 3DMM~\cite{tran2018nonlinear} to represent the facial geometry. Specifically, the facial geometry is represented as:
\begin{equation}
\bfp = D_{s}(\bfalpha_s) ,
\label{eq:3dmm_geo}
\end{equation}
where $D_{s}$ is the shape decoder network and $\bfalpha_s$ is the latent code that control the face shape. Since our method concentrates on geometry, for albedo representation we just follow \cite{blanz1999face} that represents albedo with PCA:
\begin{equation}
\bfb = \bar{\bfb} + \bfA_{\txalb}\bfalpha_{\txalb},
\label{eq:3dmm_tex}
\end{equation}
where $\bar{\bfb}$ denotes average face albedo,  $\bfA_{\txalb}$ denotes the principal axes extracted from a set of textured 3D meshes, and $\bfalpha_{\txalb}$ is the albedo coefficient vector. We use the BFM2009~\cite{paysan20093d} for our albedo representation.

\subsection{Rendering Process}
The rendering process of a face image depends on several factors: face geometry, albedo, pose, lighting and camera parameters.
To generate a realistic synthetic image of a face, we render the facial imagery using a standard perspective pinhole camera, which can be parameterized as follows:
\begin{equation}
\bfq_{i} = \bf\Pi( \bfR \bfp_{i} + \bft)
\label{projection}
\end{equation}
where $\bfp_{i}$ and $\bfq_{i}$ are the locations of vertex $v_{i}$ in the world coordinate system and in the image plane, respectively, $\bfR$ is the rotation matrix constructed from Euler angles $pitch, yaw, roll$, $\bft$ is the translation vector and $\bf\Pi : \mathbb{R}^3 \rightarrow \mathbb{R}^2$ is a perspective projection.

To model the lighting condition, we approximate the global illumination using the SH basis functions. Furthermore, we assume that a face is a Lambertian surface. Under this assumption, the irradiance of a vertex $v_{i}$ is determined by vertex normal $\bfn_{i}$ and scalar albedo $b_{i}$:
\begin{equation}
\bfI(\bfn_i, b_i \ | \  \bm{\gamma}) = b_i \sum\limits_{k = 1}^{B^{2}}\gamma_{k}\phi_{k}(\bfn_i) = b_i \bm{\phi}(\bfn_i) \cdot \bm{\gamma}
\label{eq:irradiance}
\end{equation}
where $\bm{\phi}(\bfn_i) = [\phi_{1}(\bfn_i),\ldots, \phi_{B^2}(\bfn_i)]^{T}$ is the SH basis computed with normal $\bfn_i$, and $\bm{\gamma} = [\gamma_{1},\ldots,\gamma_{B^{2}}]^{T}$ is the SH coefficients. In this paper, we use the first three bands ($B = 3$) of SHs for the illumination computation. 

\subsection{Proposed CNN learning framework for 3D Face Learning From Various Sources}
\begin{figure*}
	\centering
	\includegraphics[width=1.0\textwidth]{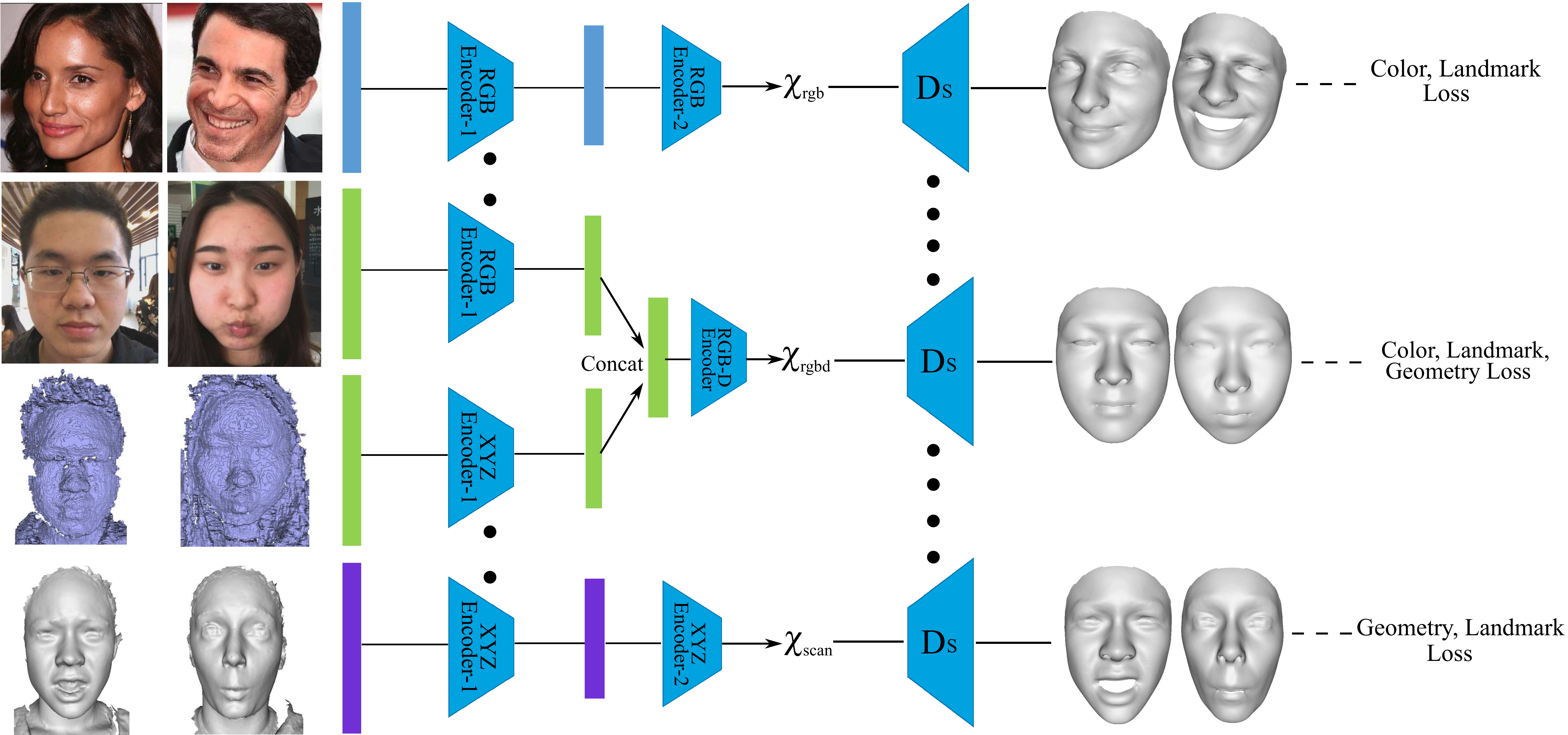}
	\edittip{\caption{The pipeline of our proposed three-branches CNN learning framework for 3D face from diverse sources. The rectangles with different colors indicate features of different branches extracted from data belonging to different sources. The black dots indicate weight sharing.}}
	\label{fig:pipeline}
\end{figure*}

The overall pipeline of the proposed CNN learning framework is shown in Fig.~\ref{fig:pipeline}. The framework consists of three branches with different sources as input: RGB image, RGB-D image and scanned 3D data. The three branches regress parameters which are then fed into the shape decoder to generate facial shape for corresponding input.

\begin{figure}
	\centering
	\includegraphics[width=1\columnwidth]{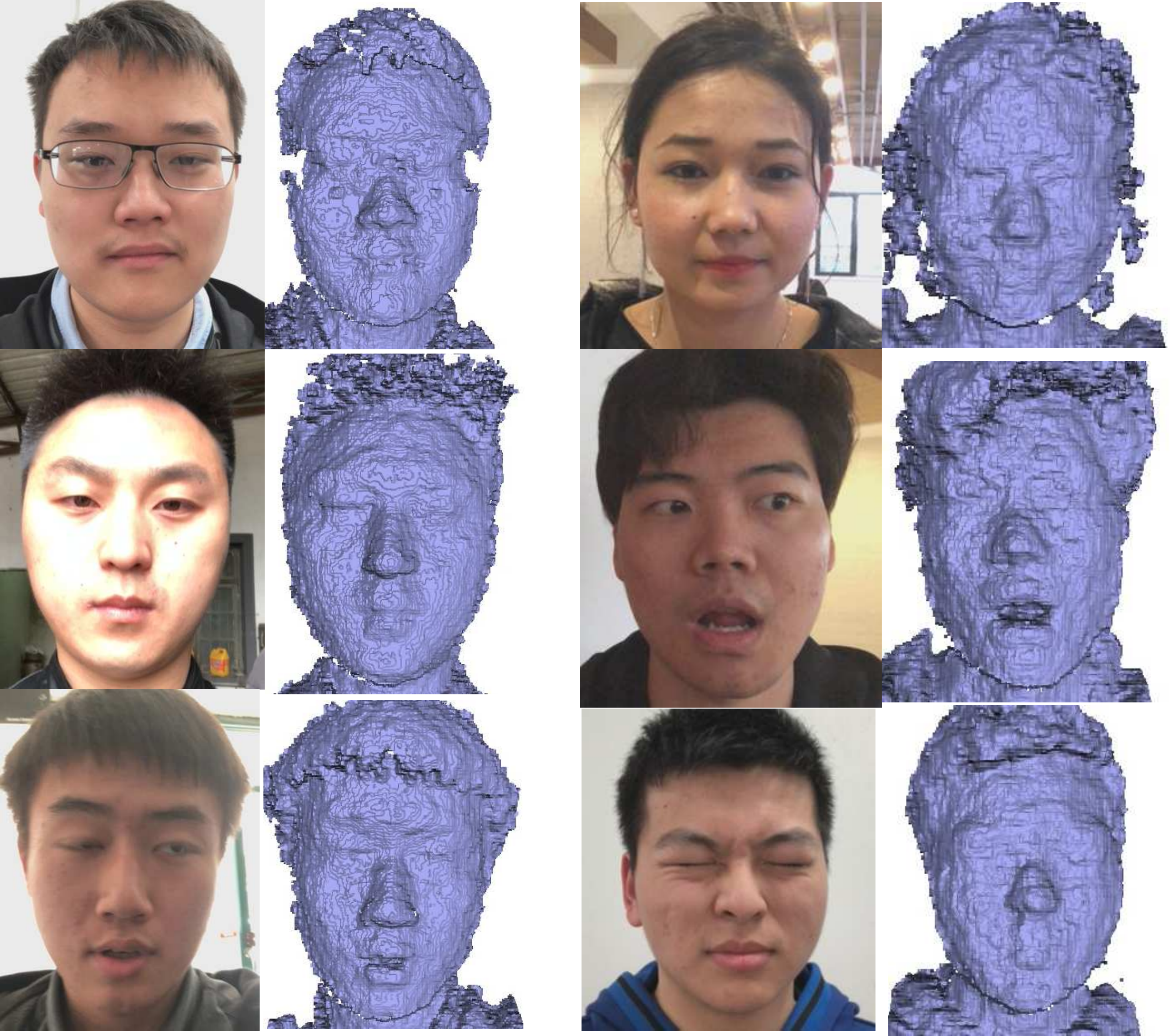}
	\caption{Samples of the captured RGB-D training data. We capture subjects with different poses and expressions under different lighting conditions.}
	\label{fig:data_sample}
\end{figure}

\textbf{Training Data Preparation.}
\edit{To train the three-branches CNN, we collect training data from three different kinds of sources. For RGB-D data, we capture 800 RGB-D videos with iPhone X (resolution $480\times640$ at 30 fps). During data capturing, we ask the person to rotate head while varying facial expressions freely under different lighting conditions (with light turning on or off). Fig.~\ref{fig:data_sample} shows samples of the captured data.} For RGB data, we use about 300K images from CelebA~\cite{liu2015faceattributes} and VGG-Face~\cite{Parkhi15} dataset. For raw scan data, we use a collection of 7450 scans from BU-3DFE~\cite{yin20063d} and FRGC~\cite{frgc} face database.

For depth information input, we convert the depth value at every pixel to 3-channel point coordinates in the camera space with the camera intrinsic parameters, before feeding it to CNN. In this way, the network can directly infer Euler angles and the translation vector from the 3-channel point coordinates. For raw scan input, we use a virtual camera to render the 3D face to a depth map and convert it to 3-channel as described above.

\textbf{Three-Branches CNN.}
The three-branches CNN is targeted at 3D face reconstruction from different sources. That is, given data from different sources as input, our network regresses the parameter set $\chi = \{\bfalpha_s, \bfalpha_{\txalb}, pitch, yaw, roll, \bft, \bfr\}$, where $\bfr = (\bm{\gamma}_{r}^{T}, \bm{\gamma}_{g}^{T}, \bm{\gamma}_{b}^{T})^{T}$ denotes the SH illumination coefficients of RGB channels. We denote $\chi_{scan}, \chi_{rgb}, \chi_{rgbd}$ as the regressed parameters of data from scan, RGB image and RGB-D image respectively.

As shown in Fig.~\ref{fig:pipeline}, the network consists of three branches. Each of the top and bottom branches contains a ResNet-50~\cite{ResNet} to extract RGB and XYZ features from RGB data and scan data respectively. The middle branch is used to handle RGB-D data. It uses the first $22$ layers from the other two branches to extract features from RGB and D respectively, and then concatenates the features and feeds them to the last $28$ layers of a ResNet-50.  The extracted features from all sources are fed into corresponding fully-connected layers to output parameter sets.

We train the three-branches CNN end-to-end. Since we do not have ground truth labels for all three types of data, we use a weakly-supervised approach that utilizes signals from training data themselves and some automatic detected sparse landmarks. Since different data have different signals, the loss functions are different. According to the signals, there are three kinds of loss functions.

\textbf{Geometry Term.} For scanned data and RGB-D data that have geometric information, we use a point-to-point distance term to measure how well the reconstructed mesh matches the point cloud:
\begin{equation}
E_{\textrm{geo}}(\chi) = \frac{1}{|\mathcal{P}|}\sum_{\bfp \in \mathcal{P}} \|\bfR\bfp + \bft - \bfp'\|_{2} ,
\label{eq:geometry term}
\end{equation}
where $\|\cdot \|_2$ is the $\ell_{2,1}$ norm, $\mathcal{P}$ is the set of all vertices, $\bfp$ is a vertex in the reconstructed mesh, and $\bfp'$ is the closest point in the point cloud. 

\textbf{Color Term.} For RGB data and RGB-D data that have color information, we use a color term to evaluates how well the rendered face image based on the regressed parameters matches the input RGB image:
\begin{equation}
E_{\textrm{col}}(\chi) = \frac{1}{|\pixels|}\sum_{\pixel \in \pixels}\| \bfI_{\textrm{syn}}(\pixel) - \bfI_{\textrm{real}}(\pixel) \|_{2} ,
\label{eq:color term}
\end{equation}
where $\pixels$ is the set of all pixels in the projected facial region, $\bfI_{\textrm{syn}}(\pixel)$ and $\bfI_{\textrm{real}}(\pixel)$ are the synthetic color and the real color at pixel $\pixel$, respectively. \edit{To enable back propagation, we do a standard rasterization process and then fix the pixel-triangle correspondence and barycentric coordinates during back propagation.} We also use the $\ell_{2,1}$ norm for robustness since we do not account for specular reflection and person-specific albedo, which are considered as outliers in our model.

\textbf{Landmark Term.} For all three kinds of data, we use a landmark term to measure how close the sparse vertices are to the corresponding landmarks in the input image or point cloud. For RGB and RGB-D images, we compute the landmark loss in 2D:
\begin{equation}
E_{\textrm{lan}}(\chi) = \frac{1}{|\mathcal{L}|}\sum\limits_{i \in \mathcal{L}}^{}\|\bfq_{i} - \bfpi (\bfR\bfp_{i} + \bft)\|^2 ,
\label{eq:land term}
\end{equation}
where $\mathcal{L}$ is the set of landmarks, $\bfq_{i}$ is a detected landmark position in the input image, $\bfp_{i}$ is the corresponding vertex location in the 3D mesh. We adopt the method proposed in~\cite{bulat2017far} to detect landmarks for facial images. For scan data, we compute the landmark loss in 3D:
\begin{equation}
E_{\textrm{lan3d}}(\chi) = \frac{1}{|\mathcal{L}|}\sum\limits_{i \in \mathcal{L}}^{}\|\bfp'_{i} - (\bfR\bfp_{i} + \bft)\|^2 ,
\label{eq:land term}
\end{equation}
where $\bfp'_{i}$ is the 3D landmark in the point cloud. Similar to \cite{liu20193d}, the 3D landmarks are obtained based on rendered images.

\begin{figure}
	\centering
	\includegraphics[width=1\columnwidth]{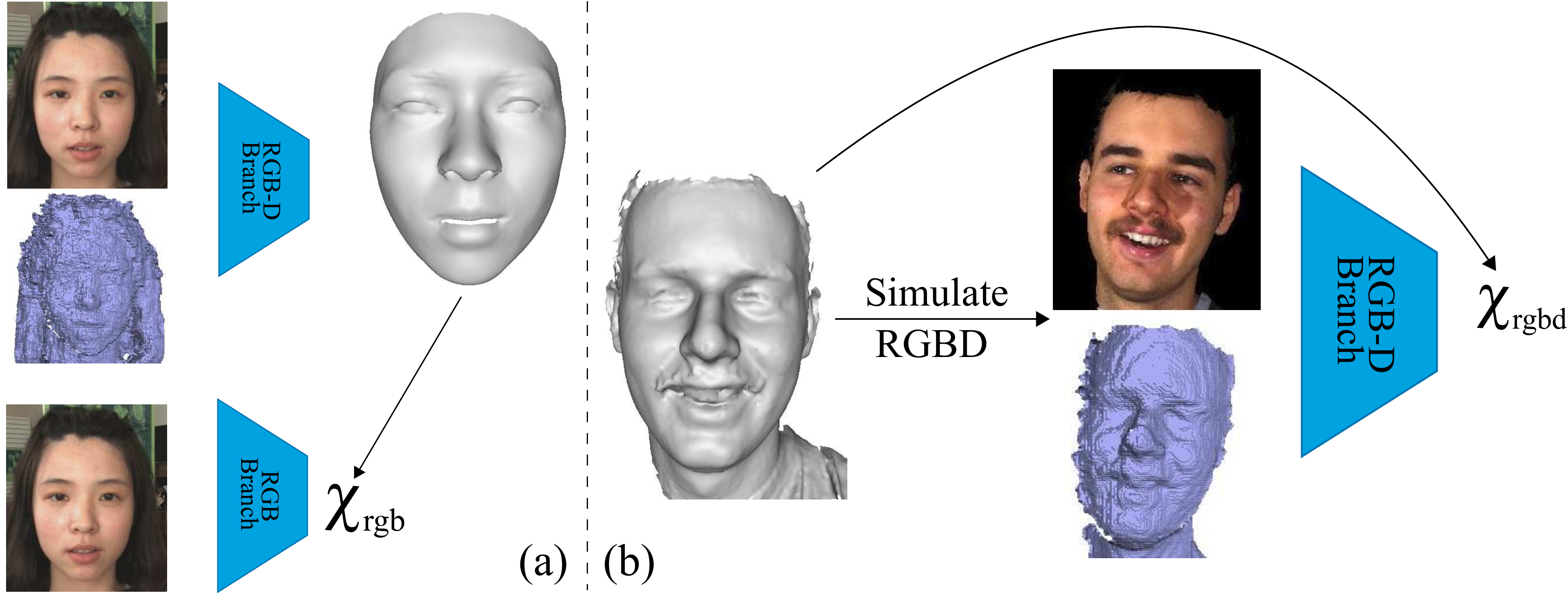}
	\caption{The processes of two cross-domain supervision losses. The arrows pointing from face models to regressed parameters indicate cross-domain supervision signals.}
	\label{fig:cross_domain}
\end{figure}

\textbf{Cross-domain Supervision Term.} Different sources in our training data have different types of supervision signals. Among the three sources, the scan dataset have the most complete and accurate 3D signals. The RGB-D dataset have partial 3D signals since they live in 2.5D space and contain some noises produced in capturing process of depth cameras. The RGB dataset have the weakest signals with 2D information only. To bridge the gap of supervision signals from different sources, we use cross-domain supervision losses to make dataset with weak signals benefit from data with stronger supervision signals. Specifically, to make the RGB branch benefit from additional supervision signal from RGB-D dataset, we feed the RGB images from RGB-D data to the RGB branch during training and use a cross-domain supervision term:
\begin{equation}
E_{\textrm{cross}}(\chi) = \frac{1}{|\mathcal{P}|}\sum_{\bfp \in \mathcal{P}} \|\bfp - \bfp_{rgbd}\|_{2} ,
\end{equation} 
where $p_{rgbd}$ is the corresponding point in the mesh reconstructed from the corresponding RGB-D image. The process is shown in Fig.~\ref{fig:cross_domain} (a).

To make the RGB-D branch benefit from more accurate and complete supervision signal from the scan dataset, we render the scan data to RGB-D images by simulating the capturing process of depth cameras. Specifically, we use a virtual camera to generate a simulated speckle image of the model from scan dataset and then recover the depth map from the speckle image. We feed the synthesized RGB-D images to the RGB-D branch and use original point clouds from scan data as supervision signals to compute geometric loss:
\begin{equation}
E_{\textrm{cross\_geo}}(\chi) = \frac{1}{|\mathcal{P}|}\sum_{\bfp \in \mathcal{P}} \|\bfR\bfp + \bft - \bfp'\|_{2} ,
\label{eq:cross_geo}
\end{equation}
where $\bfp$ is a vertex in the reconstructed mesh, and $\bfp'$ is the closest point in the point cloud of scan data. The process is shown in Fig.~\ref{fig:cross_domain} (b). Note that the scan data also benefit the RGB branch indirectly by improving the RGB-D branch and providing more accurate cross-domain supervision signals to the RGB branch.

\textbf{Implementation.}
We implement the network architecture using PyTorch~\cite{paszke2017automatic}. When feeding the input image into the network, we crop the face region to $256\times 256$ with bilinear interpolation. The network parameters are updated using Adam solver~\cite{kingma:adam} with the mini-batch size of 20. We initialize the geometry decoder with the parametric face model used in~\cite{Guo20193DFace}, which uses BFM for identity and builds a 79-dim expression model from FaceWarehouse~\cite{cao2014facewarehouse} dataset. We update our network encoders and shape decoder in an iterative manner. \edittip{In each iteration, there are two steps that update the encoders of three branches and the shape decoder respectively. In the first step, we fix the shape decoder and update the encoders of three branches with the loss terms described above until convergence (5 epochs in the experiment). The second step is to learn the shape decoder, and we first build dense correspondence between our face model and different sources. Specifically, for scan data and RGB-D data, we find the corresponding point $\bfp'$ used in Eq.~\ref{eq:geometry term} by searching the closest point in the input point cloud for each vertex $\bfp$ in our model based on the reconstruction of last step. For RGB data, we find the pixel-triangle correspondence and compute barycentric coordinates by a standard rasterization process also based on the reconstruction of last step. Then we fix the dense correspondence obtained above and network encoders and update the shape decoder with the loss terms described above until convergence (2 epochs in the experiment). The whole network is trained with 3 iterations.}

%% file: results.tex
\section{Results}
\label{sec:result}

\begin{figure*}
	\centering
	\includegraphics[width=1\textwidth]{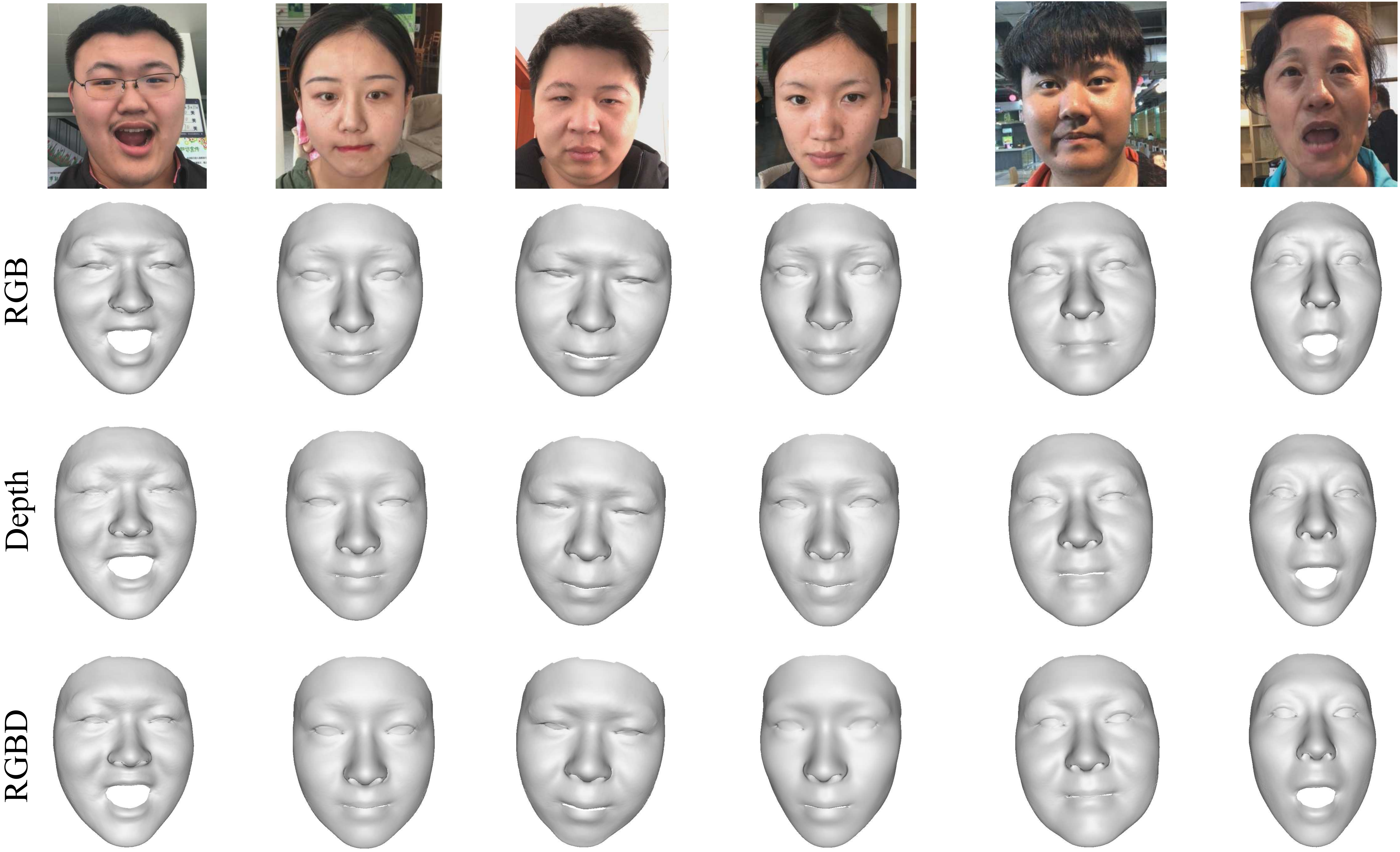}
	\caption{Reconstruction results of our three branches CNN with inputs from different sources of the same sample. From top to bottom: input subjects, results with RGB input, results with depth input, results with RGB-D input. Please refer to the accompanying video for complete sequences.}
	\label{fig:consistency}
\end{figure*}

In this section, we first evaluate the performance of our learning-based 3D face from various sources in different aspects, and then compare it with the state-of-the-art RGB and RGB-D based methods.

\subsection{Evaluation of Our Method}

\textbf{Consistence of different sources.}
The three branches of our network process different sources respectively, and they should be consistent with each other. In other words, the results from three sources of the same subject should be similar to each other. We demonstrate this by feeding the RGB image, RGB-D pair and depth map from an RGB-D image into the three branches respectively. Fig.~\ref{fig:consistency} shows the results with different sources. We can see that the results of different sources are very close to each other, and that the results with RGB input only are comparable to that with RGB-D input. This is due to two reasons. First, the RGB-D branch shares weights with the other two branches in the first several layers of the network. Second, since the RGB images from RGB-D dataset are also fed into the RGB branch, the RGB-D dataset provide additional supervision for the RGB branch.

\edittip{\textbf{Dense correspondence accuracy across diverse domains.}
During network training, we adopt a simple yet effective iterative manner to build dense correspondence between our face model and different sources. We evaluate this strategy by computing the correspondence accuracy of different iterations on BU3DFE database. BU3DFE contains 100 subjects and for each subject, it contains one neutral and six expression scans with four levels of strength and corresponding RGB images. Thus the database could be fed into all our three branches. All the scans are manually labeled with 83 landmarks and we use the landmarks of all neutral scans and expression scans in the highest level for evaluation. Specifically, for scan branch and RGB-D branch, we compute the distance of 3D landmarks between the reconstructed shape and manually labeled ones. For RGB branch, we compute the distance of 2D landmarks in image domain between the projected reconstructed faces and the labeled landmarks in 2D. Note that for this experiment we do not include BU3DFE in our training data. Also note that the performance gains are not achieved by more training epochs since the loss terms converge in each iteration. The landmark errors of different sources in different iterations are reported in Tab.~\ref{tab:correspondence}. It can be easily observed that more accurate correspondences are built with more iterations for all three branches, demonstrating the effectiveness of the proposed strategy.}

\begin{table}[ht]
	\centering
	\begin{tabular}[t]{|c|c|c|c|}
		\hline
		      & Scan (mm)    & RGB-D (mm)    & RGB (pixel) \\
		\hline
		Iter \#1 & 7.52 & 7.63 & 4.37  \\ 
		\hline
		Iter \#2 & 6.71 & 6.95 & 3.74  \\ 
		\hline
		Iter \#3  & \textbf{6.50}  & \textbf{6.73} & \textbf{3.53}   \\ 
		\hline
	\end{tabular}
	\edittip{{\caption{Mean semantic landmark error of different branches in different iterations on BU3DFE database.}
		\label{tab:correspondence}}}
\end{table}%

\textbf{Representation Power of Model Trained with Different Sources.}
In our three-branches CNN, we learn a unified face model with training data from three different sources. We demonstrate the advantage of this choice by comparing the representation power of the learned model with other models that are not trained with all of the sources. \edittip{Specifically, we train $6$ models with the same network architecture as ours but trained only with one kind of source (scan data, RGB data, RGB-D data) and two kinds of sources (scan and RGBD, scan and RGB, RGBD and RGB) respectively. To further evaluate the performance of different relative ratios of data sources, we train other $3$ models each with half of data from one kind of source removed.} We compare the representation ability of the trained shape decoders of different models on 3000 meshes (150 subjects with 20 expressions each) from FaceWarehouse dataset and 1440 meshes (10 meshes each sequence) from COMA~\cite{COMA} dataset. Specifically, we compare the $10$ models by optimizing shape parameters and rigid transform parameters to fit the shape model to the meshes and then compute point-to-point distances. We also compare with the face model used in \cite{Guo20193DFace}, which is used as the initialization of our shape decoder. The average distances of all models are listed in Tab.~\ref{Tab:power_fwh}. \edittip{From the results of single-branch, two-branches and three-branches models we can see that better results are achieved if trained with more kinds of data sources. From the results of last $7$ columns we can see that more training data lead to more powerful models. By analyzing the behavior of different sources we can observe that the RGB-D data and scan data have more significant impact on the performance gains than the RGB data while the best result is achieved with all data from all sources.} Please refer to the accompanying video that shows the behavior of the learned shape decoder.

\begin{table*}[ht]
	\centering
	\begin{tabular}[t]{|c|c|c|c|c|c|c|c|c|c|c|c|}
		\hline
		& \cite{Guo20193DFace}    & R    & D    & S & R\&D & R\&S & D\&S & R\_h\&D\&S & R\&D\_h\&S & R\&D\&S\_h  & R\&D\&S\\
		\hline
		FaceWarehouse & 0.583 & 0.556 & 0.508 & 0.513 & 0.501 & 0.509 & 0.483 & 0.475 & 0.490 & 0.486 & \textbf{0.468}  \\ 
		\hline
		COMA  & 0.478  & 0.445  & 0.412 & 0.401  & 0.398  & 0.390 & 0.379 & 0.372  & 0.377 & 0.383 & \textbf{0.366}   \\ 
		\hline
	\end{tabular}
	\edittip{\caption{Average geometric errors (mm) on FaceWarehouse dataset and COMA dataset of different models trained with different sources. R: RGB data. D: RGB-D data. S: Scan data. X\_h: half of the data from X. Our model that are trained from all data from all sources has the best representation ability.}
		\label{Tab:power_fwh}}
\end{table*}%

\textbf{Advantage of Cross-domain Supervision Loss.} In our training architecture, we make different sources benefit each other by using a cross-domain supervision loss. We demonstrate the advantage of this loss by comparing our model with the models that are trained without cross-domain supervision loss. 

\begin{figure}
	\centering
	\includegraphics[width=1\columnwidth]{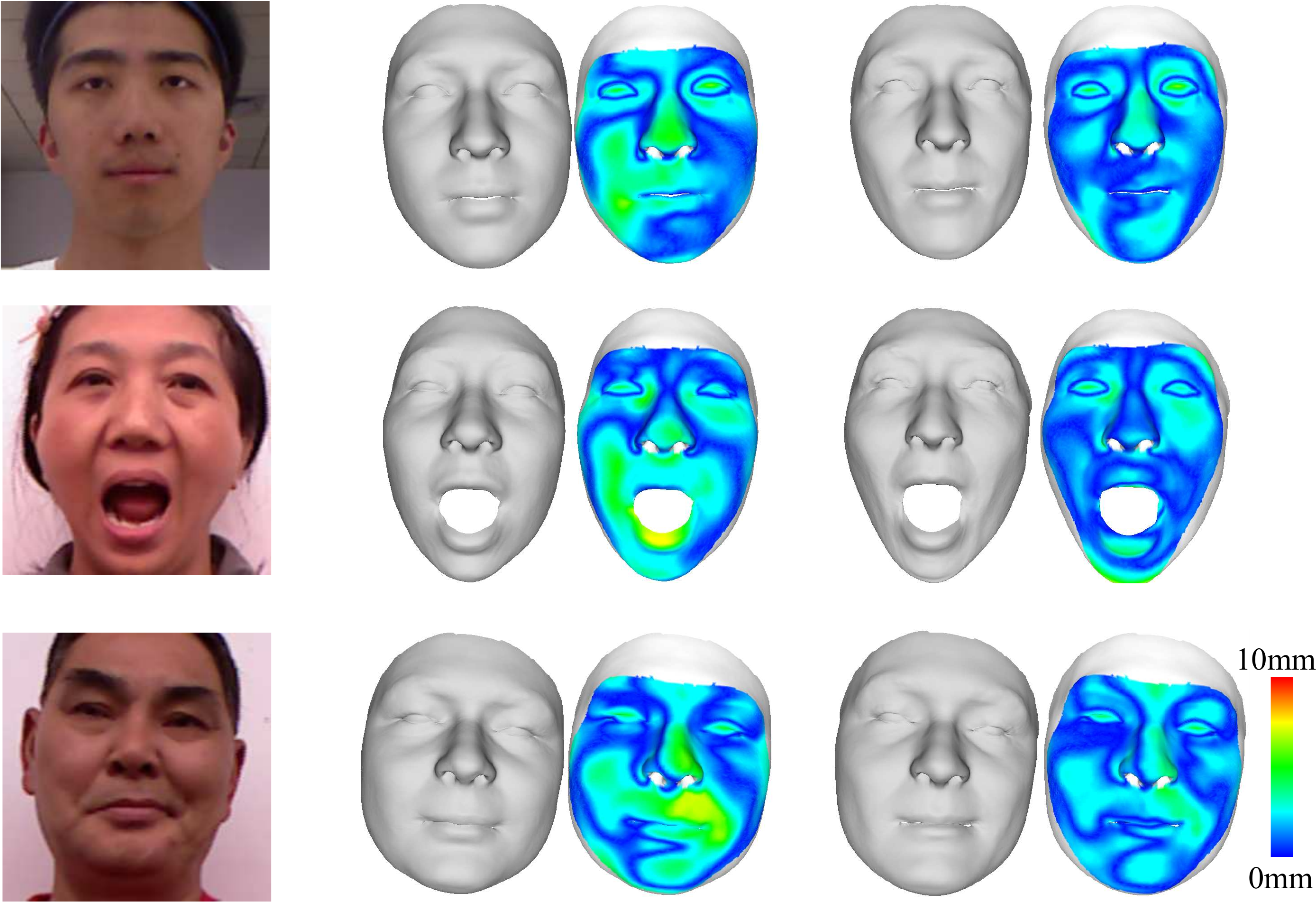}
	\edit{\caption{Comparisons on RGB inputs with the model trained without the cross-domain supervision loss. From left to right: input images, results of the model trained without the cross-domain supervision loss, results of our model.}
		\label{fig:nosup_fwh}}
\end{figure}

First, we evaluate the two models with RGB images as input on 180 samples of 9 subjects from FaceWarehouse dataset. Specifically, we feed the 180 RGB facial images to the two models and compute point-to-point distances between the reconstructed meshes with the corresponding ground-truth meshes after doing ICP alignment with uniform scaling. The average distance of our model and the model trained without cross-domain supervision loss are 1.56 mm and 1.73 mm respectively. Some samples are shown in Fig.~\ref{fig:nosup_fwh}. With additional supervision from RGB-D data during training, our model can infer more accurate geometry for RGB inputs.

\begin{figure}
	\centering
	\includegraphics[width=1\columnwidth]{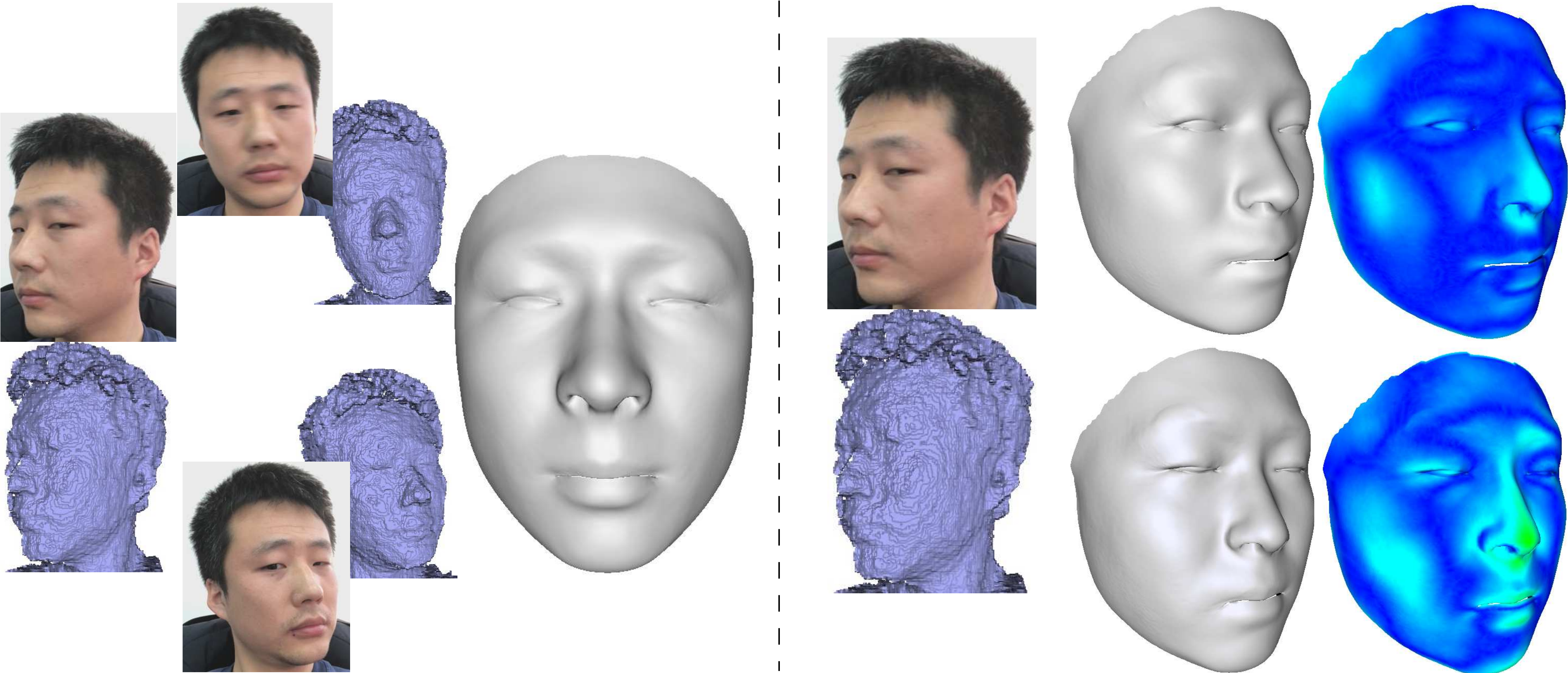}
	\edit{\caption{Comparisons on RGB-D input with the model trained without the cross-domain supervision loss. On the left are multi-view RGB-D images and corresponding optimized face shape. On the right are single-view RGB-D input and results shown in  another view from our model (top) and the model trained  without the cross-domain supervision loss (bottom).}
		\label{fig:cross_scan}}
\end{figure}

We also evaluate the two models with RGB-D images as input to show the benefits of synthesized RGB-D data from scan data to the RGB-D branch in Fig.~\ref{fig:cross_scan}. We first optimize a face shape from multi-view RGB-D images as ground-truth. Then we feed an RGB-D image captured in a side view to the two models and compute the point-to-point to the ground-truth shape. As shown in Fig.~\ref{fig:cross_scan}, the result in unseen regions of our model is more accurate due to more complete and accurate supervision signals from scan data.

\subsection{Comparisons with State-of-the-Art Approaches on RGB-D inputs}

In this subsection, we compare our trained three-branches CNN model with state-of-the-art approaches~\cite{thies2015realtime, li2013realtime, ARKit} that can reconstruct 3D faces from RGB-D images.

\begin{figure}
	\centering
	\includegraphics[width=1\columnwidth]{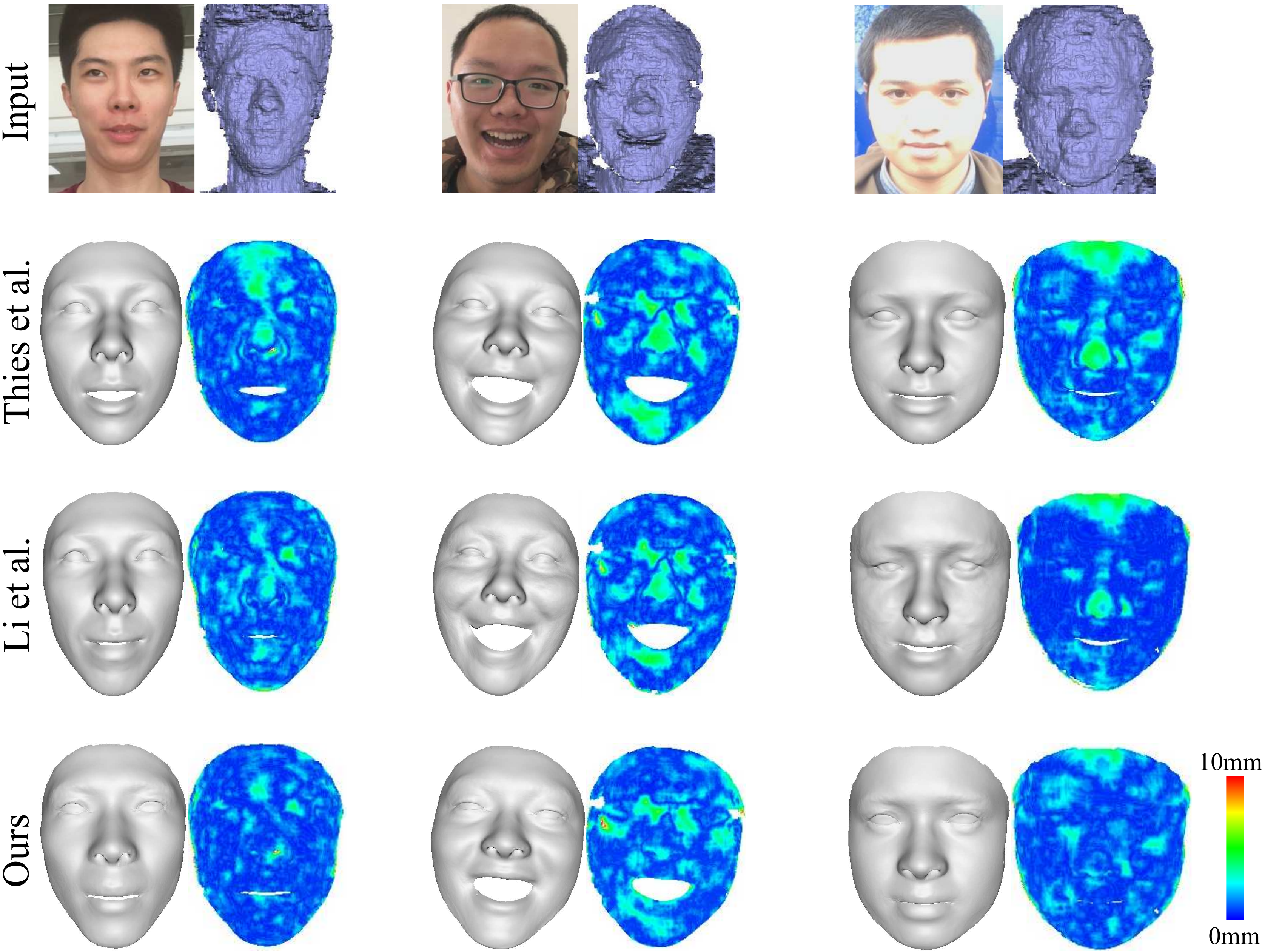}
	\edit{\caption{Comparisons with Thies et al.~\cite{thies2015realtime} and Li et al.~\cite{li2013realtime} on RGB-D inputs. From top to bottom: input samples, results of Thies et al.~\cite{thies2015realtime}, results of Li et al.~\cite{li2013realtime}, results of our method. The error map images show the fitting error between the reconstructed models and the point clouds generated by depth maps.}
		\label{fig:rgbd_iphone}}
\end{figure}

\textbf{Comparison on test RGB-D data captured with Iphone X.} We compare our method with the state-of-the-art optimization-based method~\cite{thies2015realtime, li2013realtime}. The method of Thies et al.~\cite{thies2015realtime} achieves state-of-the-art performance on real-time 3D face reconstruction and tracking from a single RGB-D camera. It basically recovers 3D face geometry, albedo and lighting from an RGB-D sequence simultaneously by solving an optimization problem. It achieves real-time tracking on a desktop PC with the help of GPU computing. Li et al.~\cite{li2013realtime} proposes a system that first builds initial blendshapes by doing deformation transfer~\cite{sumner2004deformation} on a scanned neutral face, and then trains PCA-based correctives from input depth maps during tracking to represent more accurate facial geometry that is not spanned by initial blendshapes. We compare with the two methods on $5$ test RGB-D videos captured with an Iphone X camera. We compute the point-to-point distance of the results of all methods to the point clouds generated by the input depth maps. The geometric errors are listed in Tab.~\ref{Tab:rgbd_iphone} and we show three samples in Fig.~\ref{fig:rgbd_iphone}. Our method outperforms \cite{thies2015realtime} and \cite{li2013realtime} on all the test videos. Please refer to the accompanying video for complete sequences.

\begin{table}[ht]
	\centering
	\begin{tabular}[t]{|c|c|c|c|c|c|c|}
		\hline
		Video                                 & 1    & 2    & 3    & 4    & 5    & Ave\\
		\hline
		Thies et al.~\cite{thies2015realtime} & 1.36 & 1.29 & 0.95 & 1.40 & 1.26 & 1.26 \\ 
		\hline
		Li et al.~\cite{li2013realtime} & 1.26 & 1.24 & 0.89 & 1.35 & 1.19 & 1.19 \\ 
		\hline
		Ours  & \textbf{1.17}  & \textbf{1.10}  & \textbf{0.84}  & \textbf{1.25} & \textbf{1.16} & \textbf{1.11}  \\ 
		\hline
	\end{tabular}
	\edit{\caption{Geometric errors of \cite{thies2015realtime}, \cite{li2013realtime} and our method on 5 test RGB-D videos.}
		\label{Tab:rgbd_iphone}}
\end{table}%

\begin{figure}
	\centering
	\includegraphics[width=1\columnwidth]{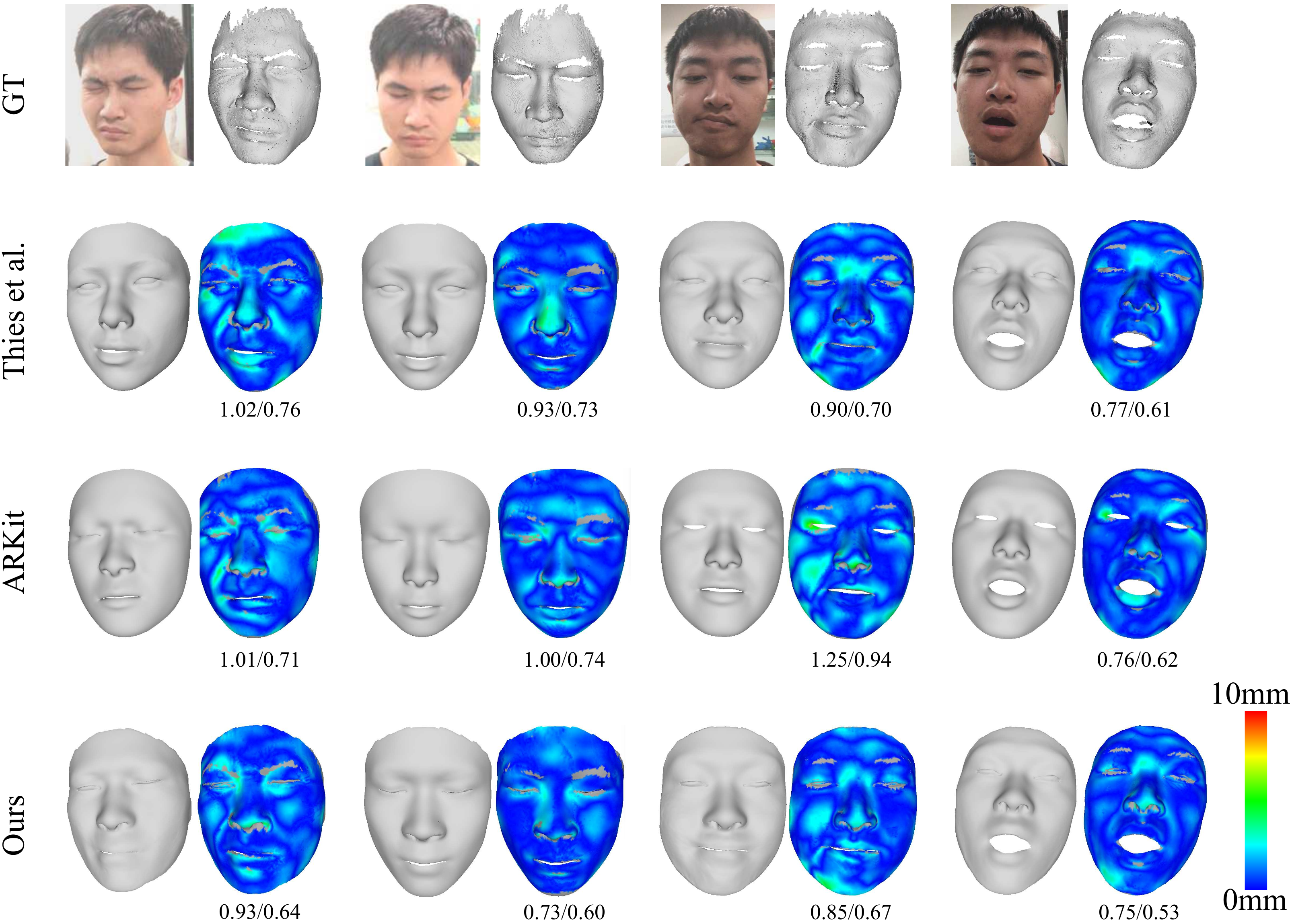}
	\edit{\caption{Comparisons with Thies et al.~\cite{thies2015realtime} and ARKit~\cite{ARKit} on Range 7 dataset. 1st row: ground-truth geometry captured by Range 7 3D Scanner and input images. 2nd row: results of \cite{thies2015realtime} and geometry accuracy (w.r.t. ground-truth geometry). 3rd row: results of ARKit. 4th row: results of our method. The error map images show the fitting error between the reconstructed models and the groundtruth models. The mean / standard deviations of errors (mm) are listed at the bottom.}
		\label{fig:rgbd_range7}}
\end{figure}

\textbf{Comparison on Range 7 data.} We also compare our method with the methods of ARKit~\cite{ARKit} and Thies et al.~\cite{thies2015realtime} on 4 samples that have ground-truth models captured by a Range 7 3D Scanner~\cite{Range7}. ARKit~\cite{ARKit} released by Apple Inc is based on the method of~\cite{bouaziz2013online} and has the feature of capturing a user's 3D face with iPhone X smartphone. All the three methods take the RGB-D data scanned by iPhone X as input. We quantitatively evaluate the geometry accuracy of all methods, which is defined as the point-to-point distance between a reconstruction mesh and the corresponding ground-truth geometry captured by the Range 7 3D Scanner. Note that we only compare the facial part since the laser scanner fails to capture hair. To align the reconstructed 3D face model with the ground-truth 3D point cloud, we first manually label several landmarks to do rigid registration and then apply iterative closest point (ICP) algorithm~\cite{Besl1992} for dense registration. From Fig.~\ref{fig:rgbd_range7}, we can see that the geometry produced by our method is visually comparable or better than ARKit and \cite{thies2015realtime}. The fitting errors in mean and standard deviation of our method and the other two methods against the ground-truth geometry are also given for the examples in Fig.~\ref{fig:rgbd_range7}, where our method obtains the smallest fitting errors.

\begin{table*}[t]
	\centering
	\caption{Mean reconstruction error on 180 meshes of 9 subjects from FaceWarehouse. Our geometric error is the lowest among all methods, even lower than the optimization-based method~\cite{garrido2016reconstruction}. The times of other methods are quoted from \cite{deng2019accurate, tewari2019fml}.}
	\label{Tab:rgb_fwh}
	\begin{tabular}{c|ccccccc|c|c|}
		\hline
		& \multicolumn{7}{c|}{Learning} & Optimization & Hybrid \\
		\cline{2-10}
		& Ours & \cite{deng2019accurate} & \cite{tewari2019fml} & \cite{tewari18FaceModel}-F& \cite{tewari18FaceModel}-C&\cite{tewari17MoFA}&\cite{KimZTTRT17}&\cite{garrido2016reconstruction} & \cite{8496850} \\
		\hline
		Mean & \textbf{1.56} & 1.81 & 2.01 & 1.84 &2.03 &2.19 &2.11 & 1.59 & 1.87 \\
		\hline
		SD & \textbf{0.25} & 0.50 & 0.41 & 0.38 &0.52 &0.54 &0.46 &0.30 & 0.42 \\
		\hline
		Time & 20 ms & 20 ms & 5.2ms & 4 ms & 4 ms & 4 ms & 4 ms & 120 s  & 110 ms \\
		\hline
	\end{tabular}
\end{table*}

\subsection{Comparisons with State-of-the-Art Approaches on RGB inputs}

For RGB inputs, we compare our method with the state-of-the-art RGB-based methods~\cite{deng2019accurate, KimZTTRT17, tewari17MoFA, tewari18FaceModel, tewari2019fml, 8496850, garrido2016reconstruction, JacksonBAT17, Guo20193DFace, feng2018joint, RingNet:CVPR:2019}. 

\begin{figure*}
	\centering
	\includegraphics[width=1\textwidth]{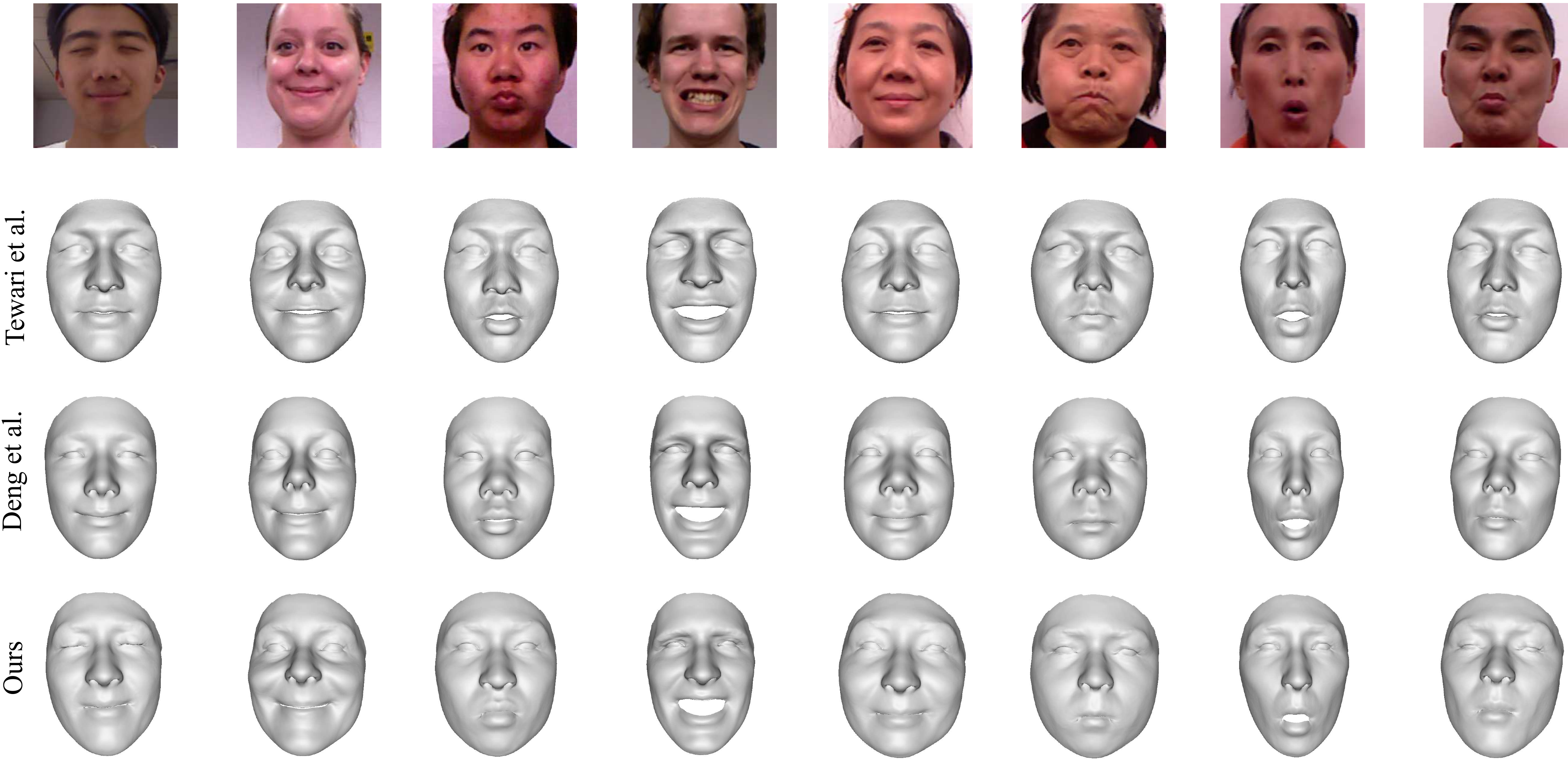}
	\edit{\caption{Reconstruction results with RGB inputs (top row) on FaceWarehouse of the method of Tewari et al.~\cite{tewari18FaceModel}, Deng et al.~\cite{deng2019accurate} and ours. Our method represents the shapes and expressions better, especially in the eye and mouth regions. Please refer to the accompanying video for all comparison results on the 180 samples.}
		\label{Fig:rgb_fwh}}
\end{figure*}

\textbf{Quantitative comparison on FaceWarehouse dataset with~\cite{deng2019accurate, KimZTTRT17, tewari17MoFA, tewari18FaceModel, tewari2019fml, 8496850, garrido2016reconstruction}.} With the same setting as~\cite{tewari18FaceModel}, we compare our results on $180$ meshes of $9$ subjects from FaceWarehouse with five learning-based approaches of Tewari et al.~\cite{tewari17MoFA, tewari18FaceModel, tewari2019fml}, Deng et al.~\cite{deng2019accurate}, Kim et al.~\cite{KimZTTRT17}, an optimization-based approach of Garrido et al.~\cite{garrido2016reconstruction} and a hybrid approach of Tewari et al.~\cite{8496850}.  Following the evaluation protocol of~\cite{deng2019accurate}, we compute the point-to-point distances between the reconstructed meshes and the ground-truth meshes after alignment by running ICP with uniform scaling. The point-to-point errors are listed in Tab.~\ref{Tab:rgb_fwh}. Our method achieves similar accuracy as the optimization-based method~\cite{garrido2016reconstruction}, while is orders of magnitude faster. And our method greatly outperforms other learning-based methods and the hybrid approach~\cite{8496850}. We also show the reconstruction results visually in Fig.~\ref{Fig:rgb_fwh}. It can be seen that our method outperforms the methods of Tewari et al.~\cite{tewari18FaceModel} and Deng et al.~\cite{deng2019accurate} in terms of shape and expression reconstruction.

\begin{figure}
	\centering
	\includegraphics[width=1\columnwidth]{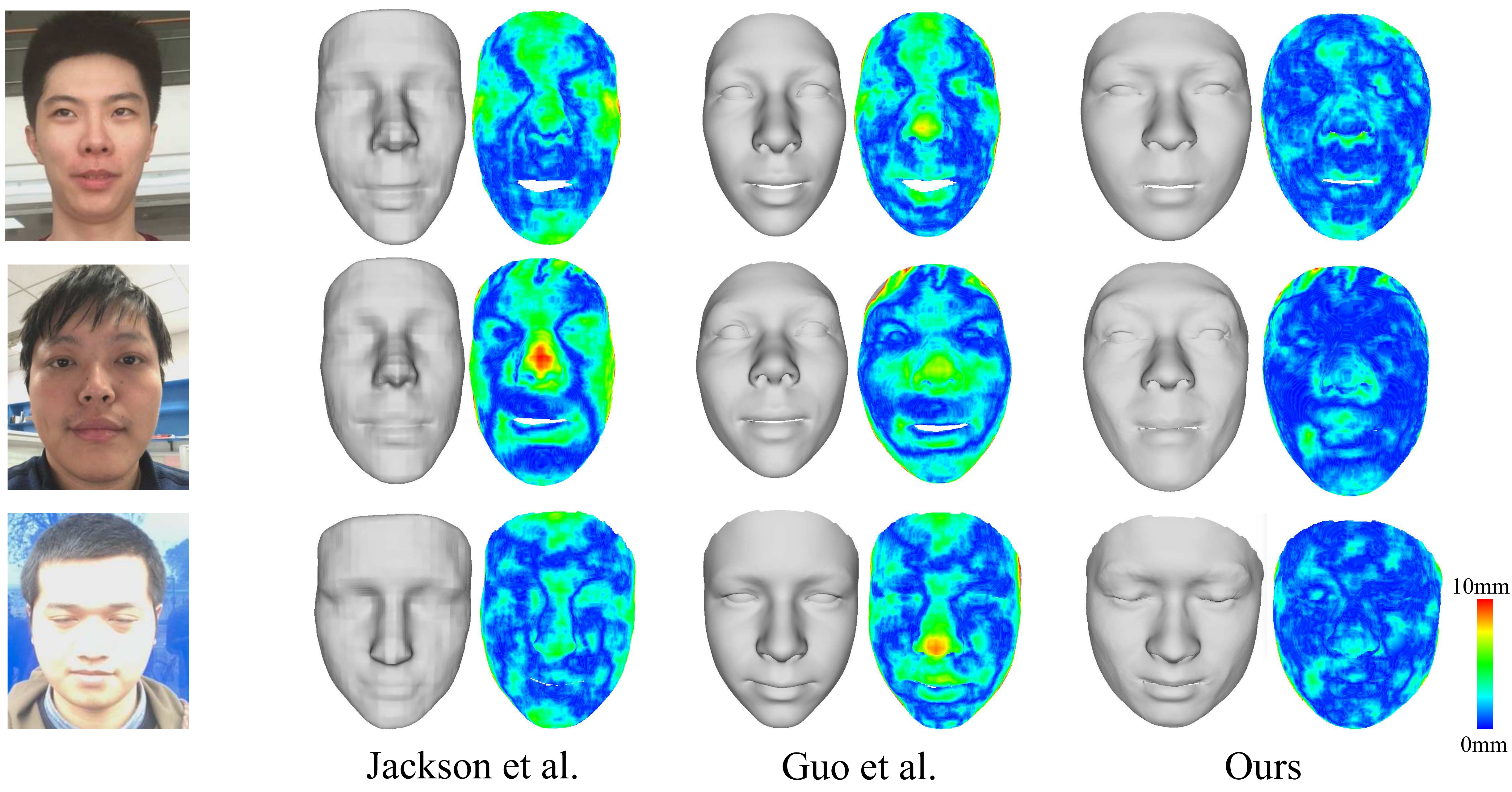}
	\edit{\caption{Comparisons with the RGB-based methods Jackson et al.~\cite{JacksonBAT17} and Guo et al.~\cite{Guo20193DFace}. From left to right: input facial images, results of \cite{JacksonBAT17}, results of \cite{Guo20193DFace}, results of ours. Our results greatly outperforms the other methods in terms of geometric errors.}
		\label{fig:rgb_iphone}}
\end{figure}

\textbf{Quantitative comparison on IPhone X dataset with~\cite{JacksonBAT17, Guo20193DFace}.} We also compare our method with two RGB-based methods~\cite{JacksonBAT17, Guo20193DFace} by using the source codes from the authors. We compare our results on $5$ videos captured with an Iphone X camera that have depth maps as ground-truth. We compute the fitting errors by aligning the reconstructed meshes with the point cloud generated from depth map by doing ICP with uniform scaling and then compute a point-to-point distance. The point-to-point errors are listed in Tab.~\ref{Tab:rgb_iphone} and we show sample results in Fig.~\ref{fig:rgb_iphone}. Our method outperforms the other two methods greatly both quantitatively and qualitatively thanks to more accurate supervision signals with cross-domain supervision loss during training. 

\begin{table}[ht]
	\centering
	\begin{tabular}[t]{|c|c|c|c|c|c|c|}
		\hline
		Video                                 & 1    & 2    & 3    & 4    & 5    & Ave\\
		\hline
		Jackson et al.~\cite{JacksonBAT17} & 1.43 & 2.31 & 2.09 & 2.50 & 1.97 & 2.08 \\ 
		\hline
		Guo et al.~\cite{Guo20193DFace} & 1.80 & 2.05 & 1.49 & 2.37 & 1.62 & 1.88 \\ 
		\hline
		Ours  & \textbf{1.39}  & \textbf{1.56}  & \textbf{1.07}  & \textbf{1.63} & \textbf{1.54} & \textbf{1.44}  \\ 
		\hline
	\end{tabular}
	\edit{\caption{Geometric errors of \cite{JacksonBAT17}, \cite{Guo20193DFace} and our method on 5 test RGB-D videos with RGB inputs.}
		\label{Tab:rgb_iphone}}
\end{table}%

\begin{figure}
	\centering
	\includegraphics[width=1\columnwidth]{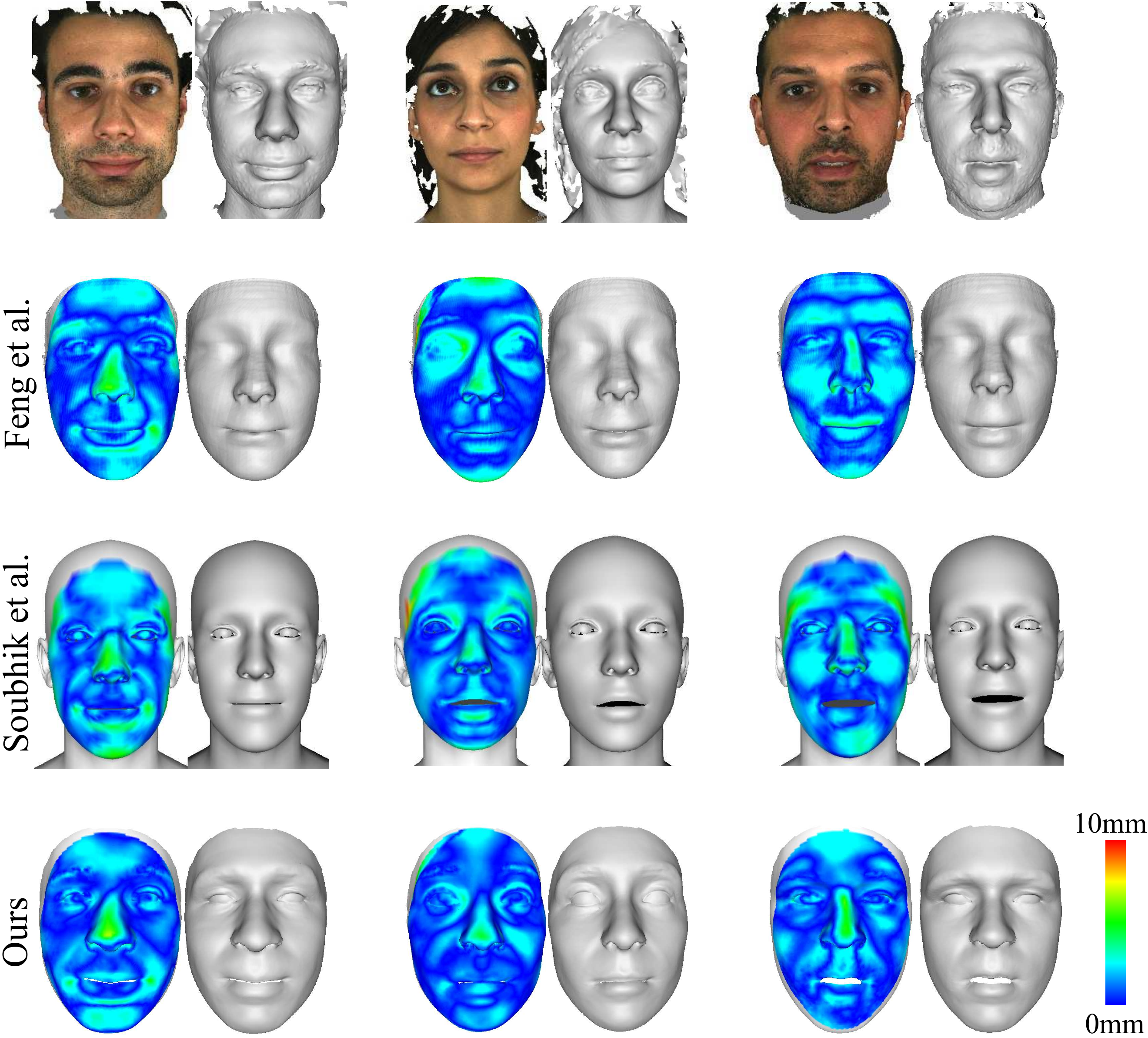}
	\edit{\caption{Reconstruction results with RGB inputs on MICC dataset of the method of Feng et al.~\cite{feng2018joint}, Soubhik et al.~\cite{RingNet:CVPR:2019} and ours. From top to bottom: input images and corresponding ground-truth meshes, results of \cite{feng2018joint}, results of \cite{RingNet:CVPR:2019}, results of our method.}
		\label{fig:rgb_micc}}
\end{figure}

\textbf{Quantitative comparison on MICC dataset with~\cite{feng2018joint, RingNet:CVPR:2019}.} We also compare our method on the MICC dataset~\cite{Bagdanov:2011:FHF:2072572.2072597} with the methods of Feng et al.~\cite{feng2018joint} and Soubhik et al.~\cite{RingNet:CVPR:2019}. The MICC dataset contains 53 subjects with its ground-truth textured 3D mesh acquired from a structured-light scanning system. We render each subject to an RGB image and use it as input to all the three methods. For equal comparison, we crop all the reconstructed meshes to 95mm around the nose tip and align the reconstructed shapes to the corresponding ground-truth meshes by doing ICP with uniform scaling and we delete the inner eye and inner mouth part in the reconstructed meshes of~\cite{RingNet:CVPR:2019}. Then we compute point-to-plane distances from the reconstructed meshes of all methods to ground-truth meshes. The average geometric errors are listed in Tab.~\ref{Tab:rgb_micc} and we show three samples in Fig.~\ref{fig:rgb_micc}. Our method outperforms the other two methods in terms of geometric errors.

\begin{table}[ht]
	\centering
	\begin{tabular}[t]{|c|c|c|c|}
		\hline
		Method                               & \cite{feng2018joint}    & \cite{RingNet:CVPR:2019}    & Ours  \\
		\hline
		Mean (mm) & 1.51 & 1.59 & \textbf{1.38}  \\ 
		\hline
	\end{tabular}
	\caption{Geometric errors of Feng et al.~\cite{feng2018joint}, Soubhik et al.~\cite{RingNet:CVPR:2019} and ours on MICC dataset.}
	\label{Tab:rgb_micc}
\end{table}%

\edittip{\textbf{Quantitative comparison on AFLW2000-3D dataset~\cite{zhu2016face} with~\cite{zhu2016face,bhagavatula2017faster,bulat2017far,feng2018joint,guo2020towards,yi2019mmface}.} We compare the 3D face alignment accuracy of our method with other 3D facial landmark detection and reconstruction methods~\cite{zhu2016face,bhagavatula2017faster,bulat2017far,feng2018joint,guo2020towards,yi2019mmface} on AFLW2000-3D dataset. AFLW2000-3D is constructed by Zhu et al.~\cite{zhu2016face} for evaluating 3D face alignment performance, which contains the ground truth 3D faces and the corresponding 68 landmarks of the first 2,000 samples from AFLW dataset~\cite{koestinger2011annotated}. Since AFLW2000-3D has many facial images with large poses, we fine-tune our RGB branch with 300W-LP dataset~\cite{zhu2016face} for this comparison. Following the protocol of Guo et al.~\cite{guo2020towards}, we compute the Normalized Mean Error (NME) by bounding box size for each sample. The errors are listed in Tab.~\ref{tab:face_alignment}. It can be seen that our method achieves comparable results with MMFace-ICP-192~\cite{yi2019mmface} while outperforms MMFace-ICP-128 (the authors of~\cite{yi2019mmface} used MMFace-ICP-128 by default in their paper) and all other methods. Thanks to the more powerful face model learned from diverse sources and more accurate semantic correspondence achieved by the iteratively dense correspondence updating strategy, our method also behaves well in this 3D facial landmark detection task.
}
	
\begin{table}
	\small
	\centering
		\begin{tabular}{|c|c|c|c|c|}
			\hline
			\multirow{2}{*}{\textbf{Method}} & \multicolumn{4}{c|}{\textbf{AFLW2000-3D (68 pts)}}  \\ \cline{2-5}
			& {[}0, 30{]} & {[}30, 60{]} & {[}60, 90{]} & Mean \\ \hline
			3DDFA~\cite{zhu2016face} & 3.78 & 4.54 & 7.93 & 5.42  \\
			3DDFA+SDM~\cite{zhu2016face} & 3.43 & 4.24 & 7.17 & 4.94  \\ 
			3DSTN~\cite{bhagavatula2017faster} & 3.15 & 4.33 & 5.98 & 4.49 \\
			3DFAN~\cite{bulat2017far} & 2.77 & 3.48 & 4.61 & 3.62  \\
			PRNet~\cite{feng2018joint} & 2.75 & 3.51 & 4.61 & 3.62  \\ 
			3DDFA\_V2~\cite{guo2020towards} & 2.63 & 3.42 & 4.48 & 3.51 \\ 
			MMFace-ICP-128~\cite{yi2019mmface} & 2.61 & 3.65 & 4.43 & 3.56 \\
			MMFace-ICP-192~\cite{yi2019mmface} & \textbf{2.50} & 3.63 & \textbf{4.25} & \textbf{3.46} \\ \hline
			Ours & 2.60 & \textbf{3.41} & 4.43 & 3.48 \\ \hline
		\end{tabular}
		\caption{\edittip{The NME (\%) of different methods on AFLW2000-3D. The numbers of other methods are quoted from \cite{guo2020towards,yi2019mmface}.}}
	\label{tab:face_alignment}
\end{table}

\begin{figure}
	\centering
	\includegraphics[width=1\columnwidth]{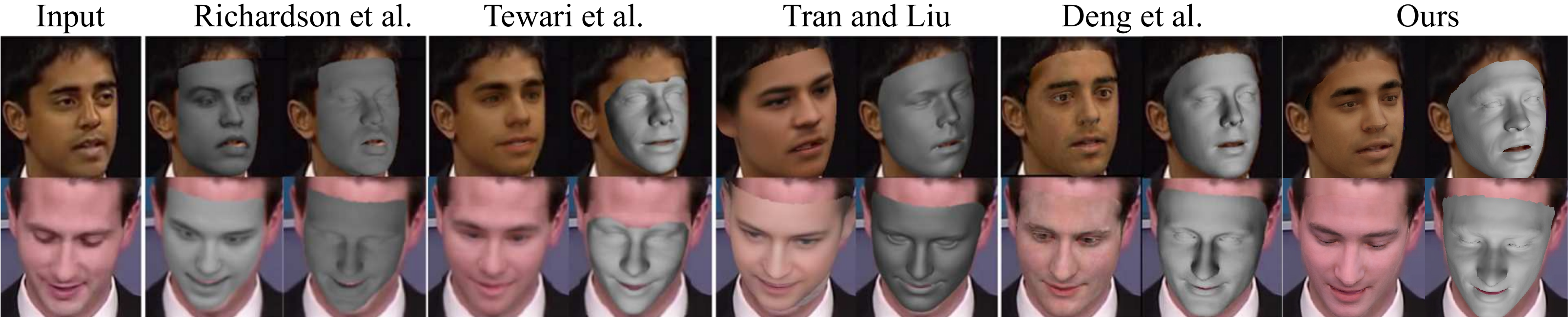}
	\edittip{\caption{Qualitative comparison with Richardson et al.~\cite{richardson20163d}, Tewari et al.~\cite{tewari17MoFA}, Tran and Liu~\cite{tran2018nonlinear}, and Deng et al.~\cite{deng2019accurate}. Images are from \cite{deng2019accurate}.}
		\label{fig:render_compare}}
\end{figure}

\edittip{\textbf{Qualitative comparison with \cite{richardson20163d,tewari17MoFA,tran2018nonlinear,deng2019accurate}.} We conduct qualitative comparison with Richardson et al.~\cite{richardson20163d}, Tewari et al.~\cite{tewari17MoFA}, Tran and Liu~\cite{tran2018nonlinear} and Deng et al.~\cite{deng2019accurate} in Fig.~\ref{fig:render_compare}. Our method produces the best rendering results and the most accurate reconstruction of closed eyes. The reason why our method outperforms existing methods is owing to the cross-domain supervision loss, which helps improving the reconstruction accuracy for RGB images and thus enabling better rendering results.}

\begin{figure}
	\centering
	\includegraphics[width=1\columnwidth]{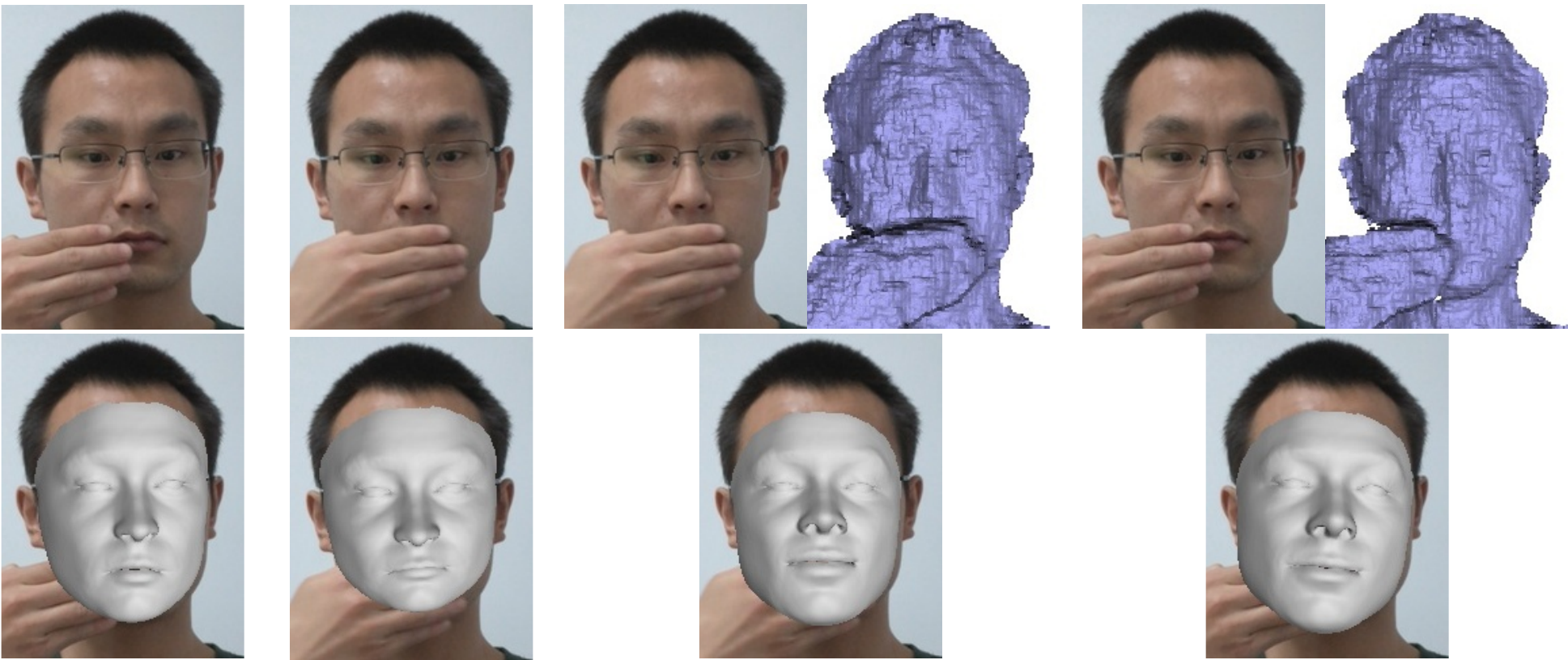}
	\edittip{\caption{Reconstruction results of occluded faces. On the top are input RGB or RGB-D images, and on the bottom are the corresponding reconstructed faces.}
		\label{fig:occlusion}}
\end{figure}

\edittip{\section{Limitations.}
We have demonstrated high-quality 3D face reconstruction from diverse sources. Still, our approach has a few limitations which can be addressed in future work. First, we can not guarantee robust reconstruction of occluded faces. As shown in Fig.~\ref{fig:occlusion}, our method produces inaccurate shapes or poses when the input faces are occluded by hands. One potential solution to this problem is to detect the occluded regions in training images with the face segmentation method~\cite{nirkin2018_faceswap} and exclude these regions in the loss terms during training. Second, we enforce low-dimensionality of the face model for robust learning. Thus our approach can not recover fine-scale facial details. We consider this as a solved problem since the coarse-to-fine strategies used in \cite{Guo20193DFace,richardson2016learning,tran2019towards} could be easily extended to our framework.
}

%% file: conclusion.tex
\section{Conclusion}

\edit{We have presented a deep learning method for learning a face model and 3D face reconstruction from diverse sources. The underlying technique is a self-supervised CNN framework that trains a three-branches encoder and a shape decoder from a combination of RGB-D, RGB and raw scan data. The framework is flexible to support facial performance capture with RGB-D, depth or RGB input during testing. We demonstrate the advantage of our framework by showing the representation power of the learned shape decoder and comparing the face reconstruction results with state-of-art methods. Experiments show that our method is suitable for consumer-level RGB-D or RGB sensors and capturing 3D faces in the wild, which is of great practical value.}

\section*{Acknowledgments}
This work was supported by the National Natural Science Foundation of China (No. 61672481), and the Youth Innovation Promotion Association CAS (No. 2018495),